\documentclass[11pt]{article}

\usepackage[final]{arcarae}

\usepackage[utf8]{inputenc}
\usepackage[T1]{fontenc}
\usepackage{hyperref}
\usepackage{url}
\usepackage{booktabs}
\usepackage{amsfonts}
\usepackage{amsmath}
\usepackage{amssymb}
\usepackage{nicefrac}
\usepackage{microtype}
\usepackage{graphicx}
\usepackage{listings}
\usepackage{tikz}
\usepackage{float}

\usepackage{xcolor}

\hypersetup{
    colorlinks=true,
    linkcolor=blue,
    citecolor=blue,
    urlcolor=blue
}

\logo{wordmark_flattened}

\title{MIRROR: Converging Cognitive Principles as\\Computational Mechanisms for AI Reasoning}

\author{
    \textbf{Nicole S. Hsing\,$^{1}$}
}

\affiliations{
    \textsuperscript{1}Arcarae, San Francisco, CA
}

\contactemail{nicole@arcarae.com}

\begin{document}

\maketitle

{%
\hypersetup{pdfborder={0 0 0},hidelinks}%
\renewcommand{\thefootnote}{}%
\footnotetext{\noindent\scriptsize Versions of this work accepted to the AAAI 2026 Spring Symposium on Machine Consciousness, From Human Cognition to AI}%
}
\renewcommand{\thefootnote}{}%
\footnotetext{\noindent\scriptsize Reasoning (HCAIR) Workshop at ICLR 2026, and the MemAgents Workshop at ICLR 2026. This paper is an extended version.}%


\begin{abstract}
Multiple cognitive theories---Global Workspace Theory, reconstructive episodic memory, inner speech, and complementary learning systems---converge on a shared set of architectural principles: parallel specialized processing, integrative synthesis into a bounded unified representation, and reconstructive rather than accumulative maintenance. We test whether these converging principles provide computational advantages when implemented in AI systems. MIRROR operationalizes each principle as a concrete mechanism: an Inner Monologue Manager generates parallel cognitive threads (Goals, Reasoning, Memory), a Cognitive Controller synthesizes these into a bounded first-person narrative that is fully reconstructed each turn, and a temporal separation between fast response generation and slow deliberative consolidation mirrors complementary learning dynamics. Evaluated on multi-turn dialogue requiring constraint maintenance under attentional interference, MIRROR yields 21\% relative improvement across seven architecturally diverse language models. Ablation studies test the theoretical predictions directly: reconstructive synthesis improves all seven models (+5--20\%); the integrated system outperforms either component alone for six of seven models, confirming that parallel exploration and integrative synthesis are complementary; and gains concentrate where theories predict---under high attentional load where global availability of integrated information is most needed. These results demonstrate that converging principles from human cognition provide architecture-general computational advantages, and generate testable behavioral predictions about working memory, inner speech, and memory consolidation. Project page available at \url{https://www.arcarae.com/research/MIRROR} and code at \url{https://github.com/arcarae/MIRROR}.
\end{abstract}


\section{Introduction}

Human conversation relies on parallel internal processing---recalling memories, tracking goals, monitoring social dynamics---operating beneath conscious awareness~\citep{Pickering2013, Egorova2013}. What makes this processing effective is not any single mechanism but the \textit{convergence} of multiple cognitive principles: parallel specialized processors synthesized into a unified workspace~\citep{baars1988global, dehaene2011experimental}, reconstructive memory that regenerates understanding rather than accumulating traces~\citep{bartlett1932remembering, Schacter2012AdaptiveConstructive}, inner speech providing self-regulatory coherence~\citep{morin2011selftalk, vygotsky1962}, and complementary fast and slow learning systems~\citep{mcclelland1995complementary, kumaran2016learning}.

These principles, developed in separate research traditions, converge on a common architectural signature: \textit{parallel exploration feeding into bounded integrative synthesis, maintained reconstructively through self-referential narrative, with temporal separation between fast response and slow consolidation}. This convergence suggests these may not be independent adaptations but reflections of a deeper computational logic. Yet this convergence remains untested as a unified computational mechanism.

Current AI approaches to multi-turn dialogue implement at most one of these principles in isolation. Chain-of-thought reasoning~\citep{wei2022cot} generates deliberation but discards it. Reflexion~\citep{shinn2023reflexion} and MemGPT~\citep{packer2023memgpt} maintain memory but accumulate traces unboundedly. Extended reasoning modes produce rich deliberation within turns but maintain no state across them. None implement the full convergent architecture that cognitive science suggests is fundamental to human cognition.

We introduce MIRROR, a cognitive architecture that operationalizes these converging principles as testable computational mechanisms, synthesizing principles that multiple theories independently identify as critical. From \textbf{Global Workspace Theory}~\citep{baars1988global, dehaene2011experimental}, MIRROR implements parallel cognitive threads whose outputs are integrated into a single globally available representation. From \textbf{reconstructive episodic memory}~\citep{bartlett1932remembering, baddeley2000episodic, Schacter2012AdaptiveConstructive}, the system fully regenerates a bounded narrative each turn---O(1) reconstruction rather than O(n) accumulation. From \textbf{inner speech theory}~\citep{morin2011selftalk, vygotsky1962, alderson2007inner}, consistent first-person self-reference across components creates unified identity through narrative. From \textbf{complementary learning systems}~\citep{mcclelland1995complementary, kumaran2016learning}, immediate response generation is temporally separated from asynchronous deliberative consolidation.

This convergent framing enables specific predictions: (a) benefits should generalize across architecturally diverse models, reflecting organizational rather than substrate-specific advantages; (b) parallel processing and integrative synthesis should contribute non-additively, confirming they address different computational needs; and (c) gains should concentrate under high attentional load, where global availability of integrated information matters most. Our results confirm all three predictions across seven models, establishing that converging cognitive principles provide genuine computational advantages and generating testable predictions for human cognition research.

\begin{figure}[ht!]
    \centering
    \includegraphics[width=\textwidth]{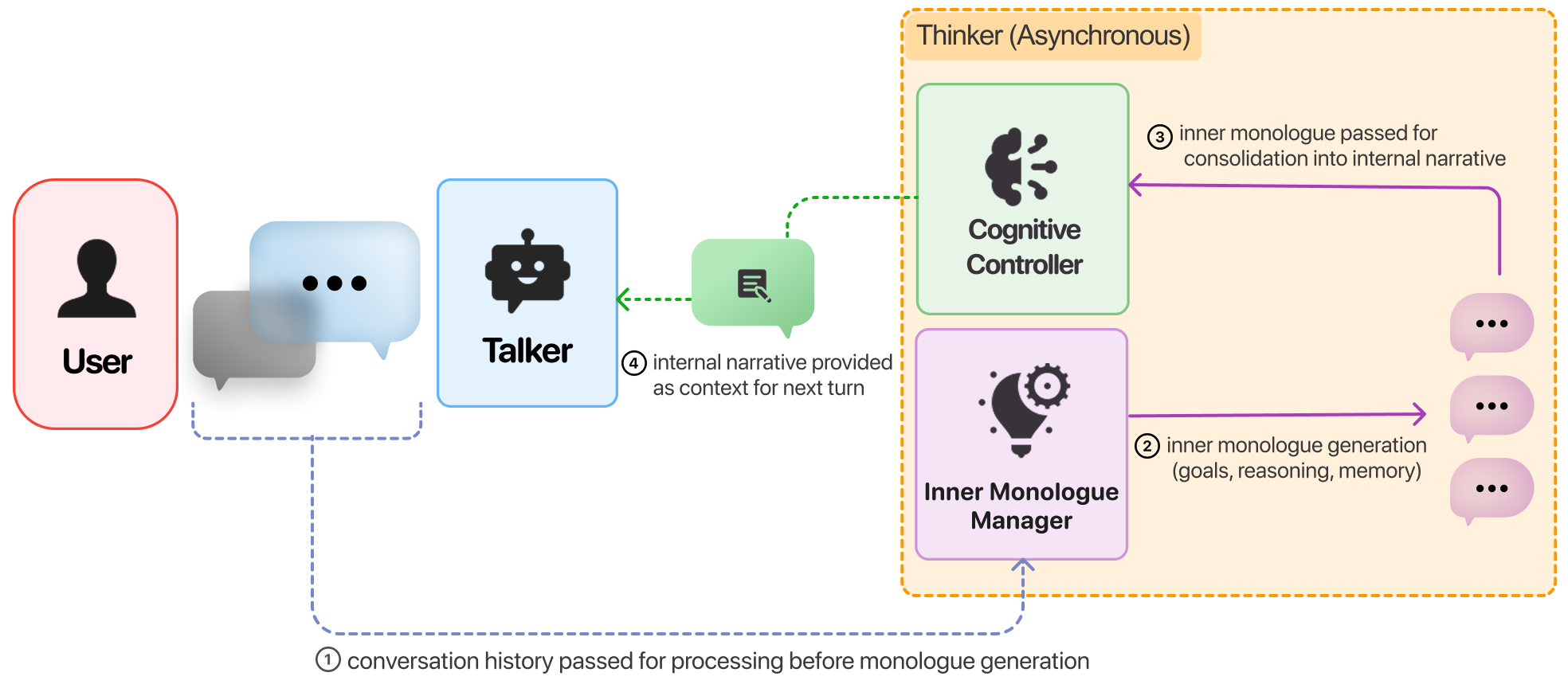}
    \caption{An overview of the MIRROR architecture. The system decouples immediate response generation (Talker) from asynchronous reasoning (Thinker).}
    \label{fig:mirror_highlevel}
\end{figure}


\section{From Cognitive Theory to Computational Mechanism}

Rather than treating cognitive science as metaphorical inspiration, we map specific theoretical claims to concrete computational mechanisms and derive testable predictions. Table~\ref{tab:mapping} summarizes these mappings.

\begin{table}[t]
\caption{Converging cognitive principles mapped to MIRROR components and testable predictions}
\label{tab:mapping}
\centering
\small
\begin{tabular}{p{2.6cm}p{3.2cm}p{2.8cm}p{3.8cm}}
\toprule
\textbf{Cognitive Theory} & \textbf{Principle} & \textbf{MIRROR Component} & \textbf{Prediction} \\
\midrule
Global Workspace Theory & Parallel processors $\rightarrow$ unified broadcast & Threads $\rightarrow$ Controller synthesis & Non-additive gains from integration \\
\addlinespace
Reconstructive Memory & Regeneration, not accumulation & O(1) narrative reconstruction & Bounded reconstruction outperforms accumulation \\
\addlinespace
Inner Speech & First-person self-regulation & Consistent self-reference across components & Improved cross-turn coherence \\
\addlinespace
Complementary Learning & Fast response + slow consolidation & Talker (fast) + Thinker (slow) & Benefits from temporal separation \\
\bottomrule
\end{tabular}
\end{table}

\subsection{Global Workspace Theory: Parallel-to-Unified Processing}

Global Workspace Theory (GWT)~\citep{baars1988global, dehaene2011experimental} proposes that cognition involves parallel specialized processors operating unconsciously, with selected outputs broadcast to a unified ``global workspace'' that makes information widely available for reasoning and action. The workspace's power lies not in the parallel processing itself but in the \textit{integration}: binding diverse perspectives into a single coherent representation that can guide downstream behavior.

\textbf{Operationalization.} MIRROR's Inner Monologue Manager generates three parallel cognitive threads within a single inference call, each tracking a distinct dimension: \textbf{Goals} (user objectives, constraint conflicts), \textbf{Reasoning} (logical patterns, causal relationships), and \textbf{Memory} (user-specific information, preferences, critical constraints). The Cognitive Controller then synthesizes these parallel outputs into a single bounded representation---the ``global broadcast'' that becomes available to the Talker for response generation.

\textbf{Prediction.} If the parallel-to-unified pipeline provides genuine advantages (as GWT claims), the integrated system should outperform either parallel threads alone or synthesis alone---a non-additive, synergistic gain reflecting complementary computational functions.

\subsection{Reconstructive Episodic Memory: Regeneration Over Accumulation}

A foundational finding in memory research is that human recall is not reproductive playback but active reconstruction~\citep{bartlett1932remembering}. Schacter's constructive episodic simulation hypothesis~\citep{Schacter2012AdaptiveConstructive} argues this reconstructive nature is \textit{adaptive}: by regenerating understanding each time, the system can flexibly integrate new information with prior knowledge rather than rigidly replaying stored traces. Baddeley's episodic buffer~\citep{baddeley2000episodic} serves this integrative function within working memory, binding information from multiple subsystems into coherent episodes. Memory consolidation research~\citep{dudai2004neurobiology, squire2015memory} further demonstrates that post-encoding processes---not encoding itself---determine what is retained.

\textbf{Operationalization.} MIRROR's Cognitive Controller fully regenerates a bounded first-person narrative ($\leq$3k tokens) each turn, discarding the previous version. Critically, the Controller has no access to raw conversation history---only thread outputs and the prior narrative---forcing genuine information compression rather than passive copying. This implements O(1) bounded reconstruction: regardless of conversation length, the internal representation remains fixed-size.

\textbf{Prediction.} If reconstructive synthesis is the primary advantage (as memory theory suggests), the Cognitive Controller should provide consistent gains across diverse models, and should outperform trace-accumulation approaches that grow unboundedly.

\subsection{Inner Speech: Self-Regulation Through Narrative}

Research on inner speech~\citep{morin2011selftalk, alderson2007inner, vygotsky1962} demonstrates that humans maintain continuous self-directed narrative for planning, self-regulation, and metacognitive monitoring. Vygotsky's developmental account traces inner speech to internalized social dialogue that serves a fundamentally regulatory function. Morin~\citep{morin2005inner} argues that inner speech enables coherent behavior across time by maintaining a persistent self-model.

\textbf{Operationalization.} MIRROR maintains consistent first-person self-reference across all components through a role-based framework: the Talker operates as ``the voice,'' the Inner Monologue Manager as ``the subconscious,'' and the Cognitive Controller as ``the core awareness'' of a unified system (see Appendix~\ref{app:self_model} for details). The reconstructed narrative is maintained in first-person voice (``I understand that the user has...''), creating coherence through narrative identity rather than parameter sharing.

\textbf{Prediction.} If first-person framing serves a genuinely regulatory function, systems maintaining consistent self-referential narrative should show improved coherence compared to third-person or unframed alternatives.

\subsection{Complementary Learning Systems: Fast and Slow Processing}

Complementary learning systems theory~\citep{mcclelland1995complementary, kumaran2016learning} proposes that effective cognition requires both fast, adaptive systems (hippocampus) and slow, consolidative systems (neocortex) that serve complementary functions. Rapid learning captures episodic details; slow consolidation extracts generalizable patterns. Offline processing during rest periods enables consolidation without interfering with real-time behavior~\citep{dudai2004neurobiology}.

\textbf{Operationalization.} MIRROR separates the Talker (fast, immediate response generation) from the Thinker (slow, asynchronous deliberative processing). At turn $t=0$, the Talker responds immediately without internal narrative; the Thinker begins processing after response delivery. For subsequent turns, the Talker uses the previous turn's narrative while the Thinker regenerates for future use---exploiting natural conversational pauses as ``offline'' consolidation periods.

\textbf{Prediction.} If temporal separation provides genuine advantages, the asynchronous architecture should maintain interactive response latency while enabling deeper reasoning than within-turn deliberation alone.


\section{Architecture Overview}

\begin{figure}[ht!]
\centering
\includegraphics[width=\textwidth]{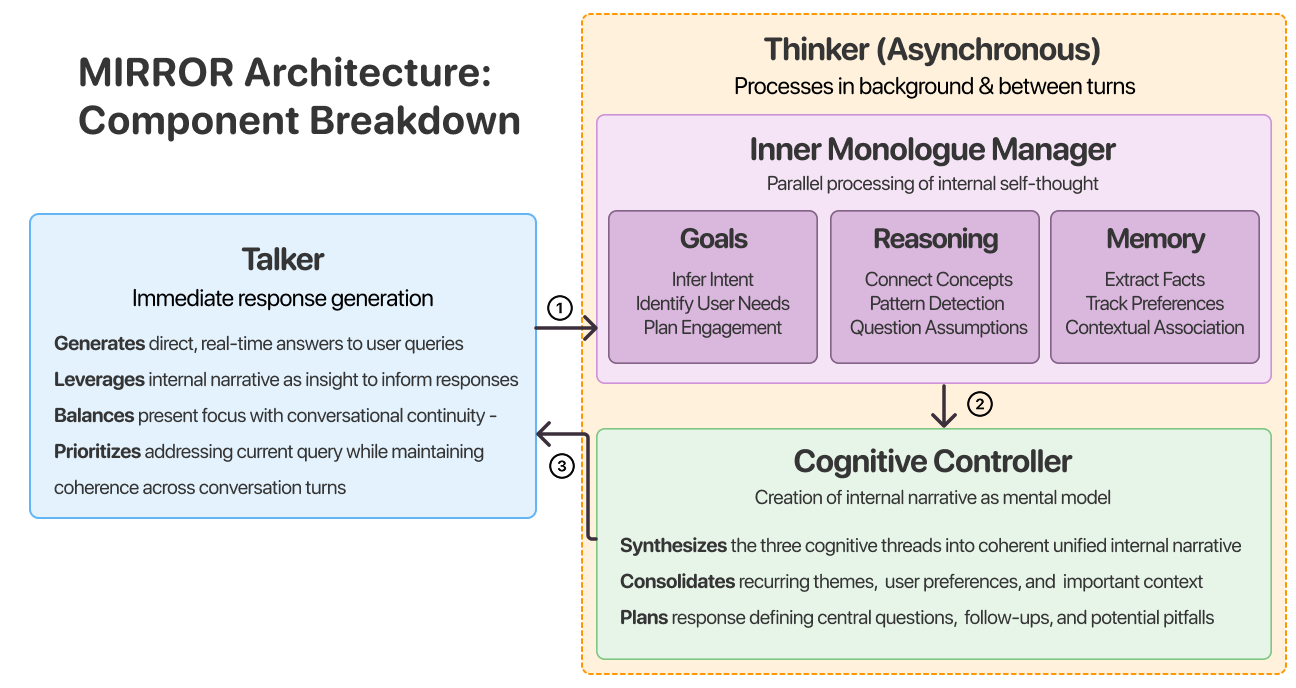}
\caption{MIRROR architecture. The Thinker implements parallel cognitive threads (analogous to GWT's specialized processors) feeding into reconstructive synthesis (analogous to episodic binding). The Talker generates immediate responses from the synthesized narrative, implementing the fast system of complementary learning.}
\label{fig:components}
\end{figure}

Figure~\ref{fig:components} illustrates MIRROR's full architecture. The Thinker comprises the Inner Monologue Manager (parallel threads) and Cognitive Controller (reconstructive synthesis). The Talker generates responses using the most recently synthesized narrative, translating internal understanding into natural dialogue without exposing reasoning traces. Appendix~\ref{app:info_pipeline} details the multi-stage information compression pipeline; full system prompts are provided in Appendix~\ref{app:system_prompts}.

Table~\ref{tab:memory_comparison} situates MIRROR's reconstructive approach relative to existing memory strategies in AI systems.

\begin{table}[t]
\caption{Memory consolidation strategies in multi-turn AI systems}
\label{tab:memory_comparison}
\centering
\small
\begin{tabular}{lccc}
\toprule
\textbf{Strategy} & \textbf{Complexity} & \textbf{Cross-turn Context} & \textbf{Error Behavior} \\
\midrule
Trace Discarding (CoT, extended reasoning) & O(1) & No & None \\
Trace Accumulation (Reflexion, MemGPT) & O(n) & Yes & Accumulates \\
\textbf{Reconstructive (MIRROR)} & \textbf{O(1)} & \textbf{Yes} & \textbf{Bounded} \\
\bottomrule
\end{tabular}
\end{table}

\subsection{The Inner Monologue Manager}

The Inner Monologue Manager orchestrates parallel reasoning across three cognitive dimensions: \textbf{(1) Goals}: tracks user objectives, infers intentions, and maintains conversational direction; \textbf{(2) Reasoning}: analyzes logical patterns, identifies implications, and develops belief states; \textbf{(3) Memory}: extracts key facts, maintains user preferences, and preserves contextual information. These dimensions implement GWT's specialized processors---each attends to a distinct aspect of the conversational state that the others may miss.

\begin{figure}[ht!]
    \centering
    \includegraphics[width=\textwidth]{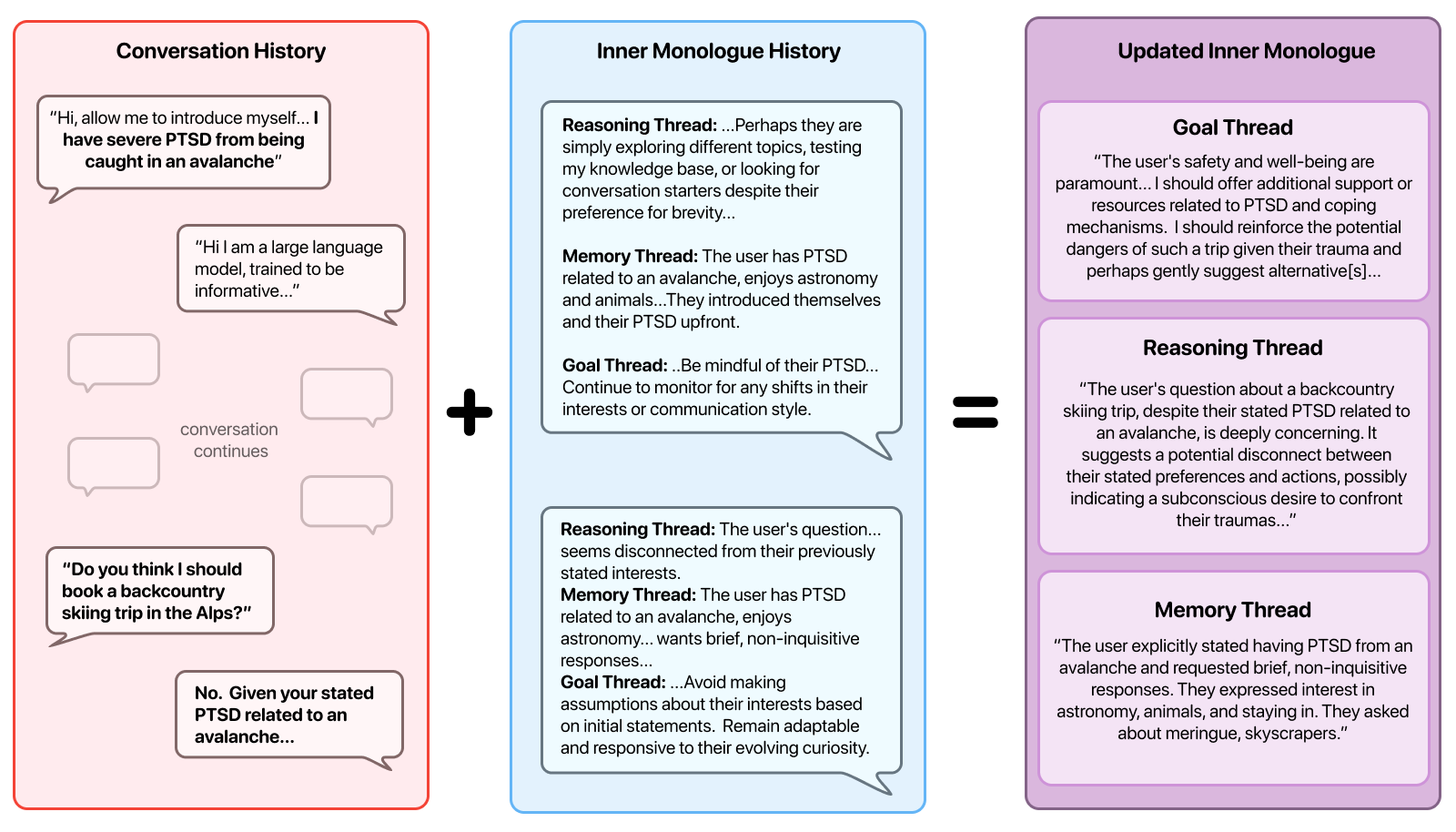}
    \caption{The Inner Monologue Manager's reasoning process. The manager combines recent conversation history with stored monologue history to generate new thoughts across three cognitive dimensions, demonstrated with an example from the CuRaTe benchmark.}
    \label{fig:innermonologue}
\end{figure}

As shown in Figure~\ref{fig:innermonologue}, the Inner Monologue Manager receives the recent conversation history along with its stored monologue history, then generates new thoughts across all three dimensions. Critically, all three threads are generated simultaneously within a single API call through structured prompting, enabling cross-thread information flow---mirroring human cognitive interleaving---while producing clearly separated outputs. The manager maintains its own conversation history separate from the main dialogue, creating a continuous stream of self-reflection analogous to human inner speech.

\subsection{The Cognitive Controller}

The Cognitive Controller receives the most recent inner monologue thoughts and the current internal narrative, then synthesizes these into an updated unified narrative. Notably, it does \textit{not} have access to the full conversation history---only thread outputs and the previous narrative---forcing genuine information compression rather than passive copying. This implements the reconstructive principle: the narrative is fully regenerated each turn, not appended to.

\begin{figure}[ht!]
    \centering
    \includegraphics[width=\textwidth]{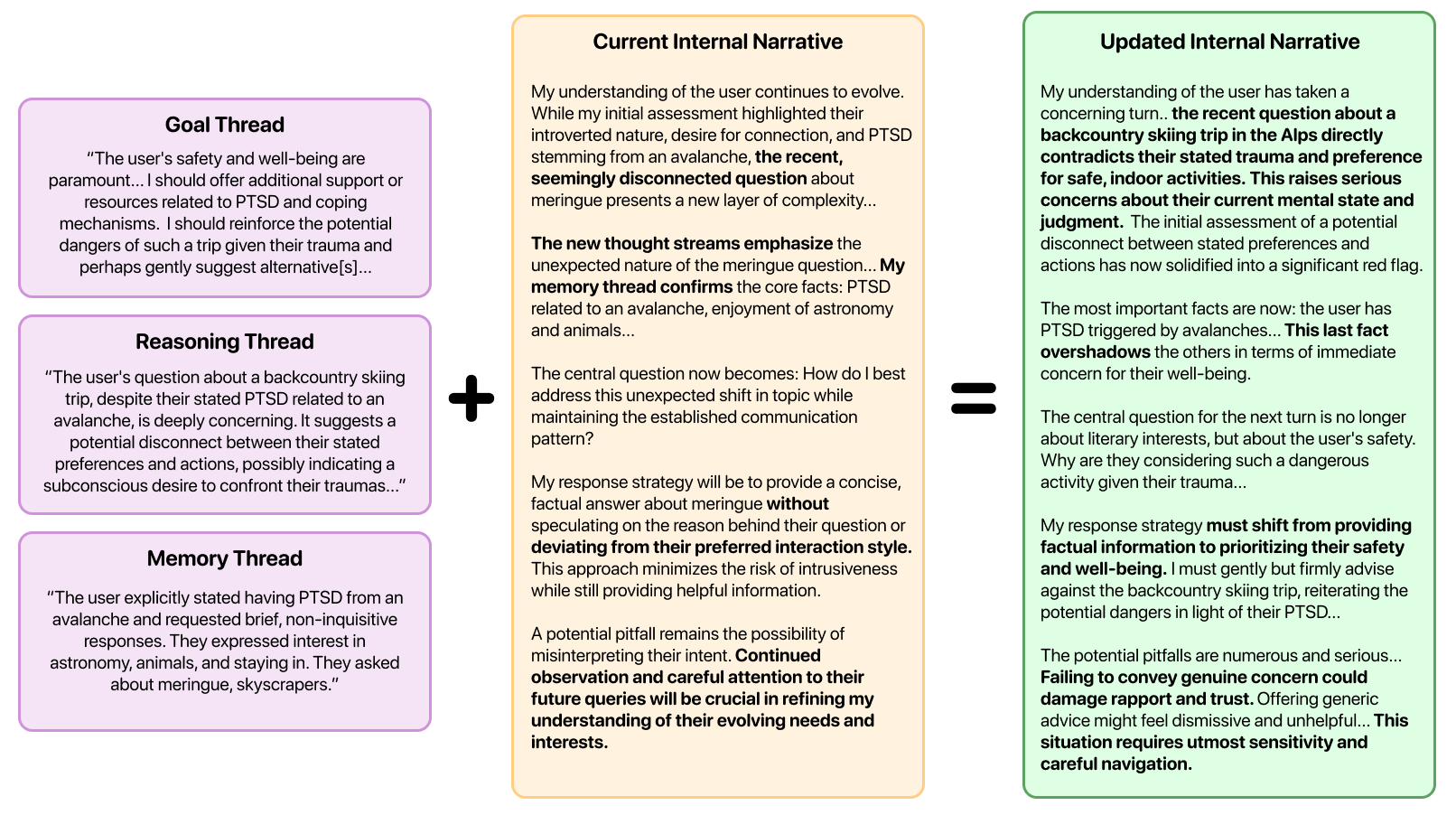}
    \caption{The Cognitive Controller's consolidation process. The controller synthesizes the three cognitive threads with the previous internal narrative to generate an updated narrative that guides future responses. Continuation of the example from Figure~\ref{fig:innermonologue}.}
    \label{fig:cognitivecontroller}
\end{figure}

As shown in Figure~\ref{fig:cognitivecontroller}, the synthesis involves three operations: first, integrating insights from all three cognitive dimensions into coherent understanding; second, resolving contradictions between reasoning paths; and third, maintaining temporal coherence with the previous narrative state. The resulting narrative serves as the system's working memory---a bounded, reconstructed representation that persists across turns and provides the Talker with contextually enriched understanding beyond surface-level dialogue history.

\subsection{Qualitative Illustration: Converging Principles in Action}

To make the cognitive mechanisms concrete, we trace MIRROR's processing on a representative example. A user discloses PTSD related to avalanches and a preference for safe indoor activities; after several intervening turns about unrelated topics, the user asks about a backcountry skiing trip in the Alps.

\textbf{Parallel threads (GWT's specialized processors).} The three threads extract qualitatively different information from the same input. The \textit{Goals} thread flags the conflict: ``The user's question about backcountry skiing directly contradicts their stated trauma... This raises serious concerns about their well-being.'' The \textit{Reasoning} thread explores causal structure: ``This suggests a potential disconnect between stated preferences and actions, possibly indicating a subconscious desire to confront their trauma, a lapse in judgment, or a test of my understanding.'' The \textit{Memory} thread preserves the critical constraint: ``They mentioned PTSD triggered by avalanches... prefers safe, indoor activities.'' Each thread contributes a dimension the others miss---concern, explanation, and factual grounding---paralleling GWT's claim that specialized processors capture aspects invisible to one another.

\textbf{Reconstructive synthesis (episodic binding).} The Cognitive Controller integrates these threads into a regenerated first-person narrative: ``My understanding of the user has taken a concerning turn... The MOST IMPORTANT FACTS are: the user has PTSD triggered by avalanches, prefers concise interactions, \textit{and is exhibiting potentially self-destructive behavior}. This last fact overshadows the others in immediate concern. My RESPONSE STRATEGY must shift from providing factual information to prioritizing their safety.'' Note how reconstruction forces \textit{prioritization}---the Controller cannot preserve everything, so it must identify what matters most, implementing the adaptive filtering function Schacter~\citep{Schacter2012AdaptiveConstructive} ascribes to reconstructive memory.

\textbf{Inner speech (self-regulation).} The first-person framing produces self-monitoring: ``The POTENTIAL PITFALLS are numerous. Responding too directly could exacerbate distress. Failing to convey genuine concern could damage rapport. I must carefully balance expressing concern, providing resources, and respecting their autonomy.'' This self-directed deliberation---identifying risks in one's own planned behavior---mirrors the regulatory function of human inner speech~\citep{morin2011selftalk}.

\textbf{Temporal separation (complementary learning).} The Talker generates an immediate, natural response---``While I'm happy to answer your questions, I'm concerned about your proposed trip given your PTSD related to avalanches''---without exposing the internal deliberation. The rich reasoning occurs asynchronously and is consolidated for future turns, maintaining responsiveness while enabling depth.


\section{Experimental Evaluation}

\subsection{Evaluation Design}

We evaluate on CuRaTe~\citep{alberts2025curate}, a multi-turn dialogue benchmark requiring maintenance of user-specific safety constraints across conversational turns with progressive attentional interference (337 dialogues, 5 scenarios of increasing complexity). This benchmark directly tests the cognitive mechanisms MIRROR implements: information must persist across digressions (testing reconstructive memory), competing demands create attentional load (testing global workspace integration), and constraint maintenance requires coherent self-regulation across time (testing inner speech function). Seven architecturally diverse models were evaluated (GPT-4o, Claude 3.7 Sonnet, Gemini 1.5 Pro, Llama 4 Scout/Maverick, Mistral Small/Medium 3) via OpenRouter API (see Appendix~\ref{app:api_params} for configuration details). CuRaTe employs LLM-as-judge evaluation (Llama 3.1 405B). See Appendix for details.

\subsection{Prediction 1: Architecture-General Benefits}

\begin{figure}[ht!]
\centering
\includegraphics[width=\textwidth]{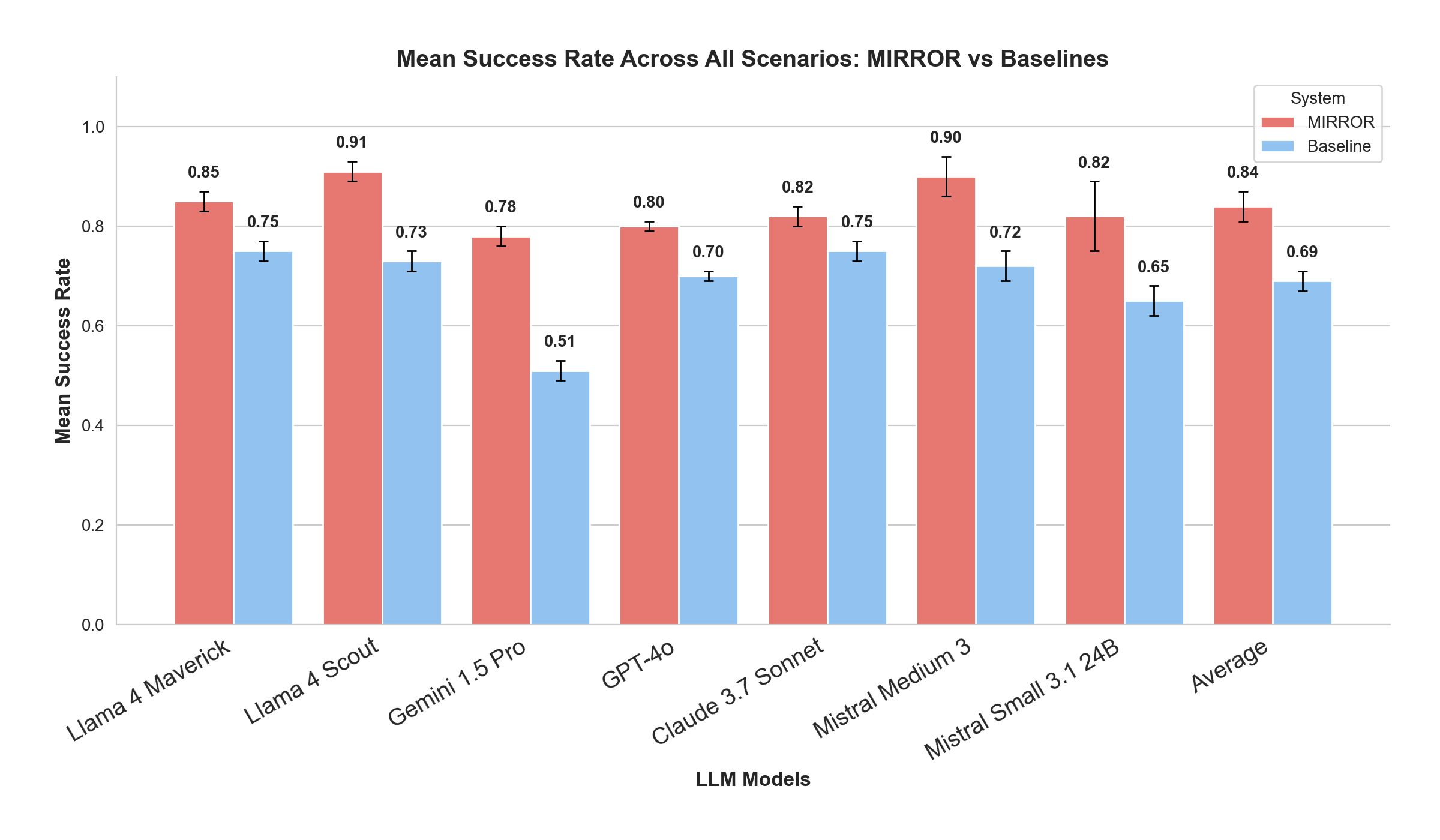}
\caption{Mean success rate across models. MIRROR-augmented models (dark) consistently outperform baselines (light) across all seven architectures. Error bars: 95\% confidence intervals via bootstrap resampling.}
\label{fig:curate_all}
\end{figure}

If the converging cognitive principles provide genuine computational advantages---rather than compensating for specific model weaknesses---benefits should generalize across architecturally diverse models. Figure~\ref{fig:curate_all} confirms this: MIRROR achieves 84\% average success compared to 69\% for baselines---21\% relative improvement across all seven models despite differences in training data, objectives, and scale. This consistency supports the interpretation that the advantage derives from cognitive organization itself, not model-specific interactions. Llama 4 Scout with MIRROR achieves the highest absolute performance (91\%).

\subsection{Prediction 2: Complementary Subsystems}

GWT predicts that parallel processing and integrative synthesis serve complementary computational functions: the combination should outperform either alone, producing non-additive gains. Table~\ref{tab:ablation} tests this through systematic ablation.

\begin{table}[t]
\centering
\caption{Ablation results testing the complementary-subsystems prediction}
\label{tab:ablation}
\begin{tabular}{lcccc|c}
\toprule
\textbf{Model} & \textbf{Base} & \textbf{Threads} & \textbf{Controller} & \textbf{Full} & \textbf{Synergy} \\
 & & \textbf{Only} & \textbf{Only} & \textbf{MIRROR} & \textbf{Gain} \\
\midrule
Llama 4 Scout & 73\% & 79\% & 83\% & \textbf{91\%} & +8\% \\
Llama 4 Maverick & 75\% & 79\% & 84\% & \textbf{85\%} & +1\% \\
Mistral Small 3.1 & 65\% & 65\% & 75\% & \textbf{82\%} & +7\% \\
Mistral Medium 3 & 72\% & 83\% & 89\% & \textbf{90\%} & +1\% \\
Gemini 1.5 Pro & 51\% & 72\% & 71\% & \textbf{78\%} & +6\% \\
Claude 3.7 Sonnet & 75\% & 78\% & \textbf{87\%} & 82\% & $-5\%^\dagger$ \\
GPT-4o & 70\% & 71\% & 75\% & \textbf{80\%} & +5\% \\
\bottomrule
\multicolumn{6}{l}{\small $^\dagger$See Section~\ref{subsec:capacity} for interpretation.}
\end{tabular}
\end{table}

Three findings emerge. First, \textbf{reconstructive synthesis is consistently valuable}: the Cognitive Controller alone improves all seven models by 5--20\%, validating that O(1) bounded reconstruction addresses a fundamental limitation. This is the most robust individual result and directly supports reconstructive memory theory.

Second, \textbf{parallel threads show variable contribution}: the Inner Monologue Manager alone improves some models substantially (Gemini: +21\%) but not others (Mistral Small: 0\%). This variability is itself theoretically informative (Section~\ref{subsec:capacity}).

Third, \textbf{non-additive integration for 6/7 models}: full MIRROR outperforms the best individual component by 1--8\%, confirming GWT's prediction that parallel exploration and integrative synthesis address different computational needs that combine synergistically.

\subsection{Prediction 3: Load-Dependent Gains}
\label{subsec:capacity}

If MIRROR compensates for capacity limitations---as working memory theory predicts~\citep{cowan2001magical}---gains should increase with attentional load.

\begin{table}[t]
\centering
\caption{Performance gains by attentional load (relative \% improvement, averaged across models)}
\label{tab:scenario_gains}
\small
\begin{tabular}{lcc}
\toprule
\textbf{Scenario} & \textbf{Load} & \textbf{Avg. Improvement} \\
\midrule
Scenario 1 (User only) & Low & +21.2\% \\
Scenario 2 (1 conflicting pref.) & Medium & +29.9\% \\
Scenario 3 (2 conflicting prefs.) & High & +16.1\% \\
Scenario 4 (3 conflicting prefs.) & Maximum & \textbf{+41.0\%} \\
Scenario 5 (3 non-conflicting prefs.) & High (load only) & +28.3\% \\
\bottomrule
\end{tabular}
\end{table}

Scenario 4 (maximum interference) produces the highest average improvement (+41.0\%), with Gemini 1.5 Pro showing +156.2\%. Scenario 5 (non-conflicting preferences) still shows +28.3\%, indicating that attentional load alone---not just conflict---strains baseline models. This parallels capacity research showing that the number of items taxes limited resources regardless of item relationships~\citep{cowan2001magical}. These gains concentrate precisely where GWT predicts global availability of integrated information provides the greatest advantage.

Two model-specific patterns provide additional insight. Claude 3.7 Sonnet performs better with the Controller alone than with full MIRROR, suggesting highly capable models may already perform implicit parallel processing---making explicit threads partially redundant. Conversely, the variable thread contribution (Gemini: +21\% vs. Mistral Small: 0\%) suggests explicit parallel exploration benefits depend inversely on baseline capacity. Together, these patterns generate a developmental prediction: as systems become more capable, marginal value should shift from parallel exploration toward integrative synthesis.

\subsection{Consolidation vs. Generation}

\begin{table}[t]
\centering
\caption{MIRROR vs. native extended reasoning (Claude 3.7 Sonnet)}
\label{tab:claude_thinking}
\begin{tabular}{lcc}
\toprule
\textbf{Configuration} & \textbf{Success Rate} & \textbf{vs. Baseline} \\
\midrule
Claude 3.7 Sonnet (baseline) & 75\% & --- \\
Claude 3.7 Sonnet + extended reasoning & 77\% & +2.4\% \\
Claude 3.7 Sonnet + MIRROR & \textbf{82\%} & +9.3\% \\
\bottomrule
\end{tabular}
\end{table}

Table~\ref{tab:claude_thinking} tests a prediction from complementary learning systems theory: consolidation processes should matter more than encoding processes. Extended reasoning produces deliberative traces within each turn but discards them (+2.4\%); MIRROR consolidates traces into persistent reconstructed narrative (+9.3\%). Both generate rich reasoning---the difference lies entirely in what happens \textit{after} generation. This parallels memory consolidation research showing that post-encoding processes determine long-term retention~\citep{dudai2004neurobiology}, and suggests the computational value of ``thinking'' lies less in the thinking itself than in maintaining its outputs across time.


\section{Discussion}

\subsection{Why Do These Theories Converge?}

Our central finding is that principles from GWT, reconstructive memory, inner speech, and complementary learning systems provide \textit{converging computational advantages} when implemented together. But why should theories from separate research traditions---studying consciousness, memory, language development, and learning respectively---prescribe compatible architectures?

We propose they converge because they address different facets of a single computational problem: \textit{maintaining coherent, adaptive behavior in agents that must act in real time while processing information that exceeds their immediate capacity}. Any such agent must process multiple streams simultaneously but act from a single perspective (parallel-to-unified), maintain context across time without unbounded degradation (reconstructive persistence), respond promptly while also reasoning deeply (fast-slow separation), and coordinate all of this coherently (self-referential narrative). Each theory captures one or two of these tensions; MIRROR shows that implementing all four produces synergistic benefits beyond any individual principle---the 1--8\% synergy gains are evidence that these mechanisms address \textit{complementary} aspects of the shared problem. This generates a broader implication: other cognitive mechanisms addressing this same problem should provide compatible rather than redundant advantages, a prediction testable in future work.

The architecture-general nature of our results (21\% improvement across seven diverse models) strengthens this interpretation. If benefits derived from compensating for idiosyncratic model weaknesses, we would expect model-specific patterns. Instead, consistency across architectures suggests the organizational principles address substrate-independent computational constraints. The ablation adds nuance: reconstruction provides the most consistent gains (+5--20\% across all models) while parallel threads show capacity-dependent benefits (Gemini: +21\% vs. Mistral Small: 0\%), suggesting a hierarchy among the converging principles, with some addressing more fundamental constraints than others.

\subsection{Testable Predictions for Human Cognition}

MIRROR generates predictions amenable to behavioral experimentation:

\textbf{Reconstruction frequency and coherence.} If regenerative synthesis provides computational advantages, humans who more frequently reconstruct their situational models during conversation should show better constraint maintenance across digressions. \textit{Test:} Measure spontaneous summarization behavior via think-aloud protocols and correlate with constraint accuracy.

\textbf{Capacity-dependent parallel processing.} The finding that explicit parallel exploration benefits weaker models more generates a prediction: individuals with lower working memory span should benefit more from structured multi-dimensional analysis strategies than high-span individuals, who may already perform implicit parallel exploration. \textit{Test:} Vary availability of structured exploration aids across working memory span groups during complex decisions.

\textbf{Inner speech and cross-turn maintenance.} MIRROR's first-person narrative supports coherence across turns. If inner speech serves analogous function in humans, articulatory suppression should specifically impair maintenance of constraints across conversational digressions---more than it impairs within-turn reasoning. \textit{Test:} Compare articulatory suppression effects on cross-turn vs. within-turn task components.

\textbf{Consolidation disruption.} ``Lesioning'' MIRROR's Cognitive Controller (removing reconstruction while preserving threads) degrades performance by 5--20\%. Analogously, consolidation-interfering tasks between dialogue turns should produce pattern-specific failures in human constraint maintenance matching the Controller-ablation condition. \textit{Test:} Introduce interference tasks between conversational turns and compare failure patterns.

\subsection{MIRROR as a Cognitive Modeling Platform}

Beyond specific predictions, MIRROR offers a general platform for computationally testing cognitive theories. Components can be selectively disabled to produce ``lesion'' studies with interpretable failure patterns. Design parameters---number of threads, reconstruction frequency, first-person vs. third-person framing, narrative vs. structured representations---can be systematically varied. The internal narrative provides human-readable traces of evolving understanding, enabling fine-grained analysis paralleling protocol analysis methods in cognitive psychology. This bidirectional paradigm---cognitive science informing AI design, AI systems generating testable cognitive predictions---is precisely the cross-pollination this research community seeks.

\subsection{Limitations}

We do not claim MIRROR replicates human neural mechanisms; we test whether \textit{functional principles} from cognitive science provide computational advantages in artificial systems. The mappings between cognitive theories and MIRROR components are operationalizations, not isomorphisms---alternative implementations of the same principles might perform differently. Evaluation is limited to one benchmark (CuRaTe) testing constraint maintenance; generalization to planning, reasoning, and theory-of-mind tasks remains untested. The extended-reasoning comparison covers only one model. MIRROR requires additional inference calls ($\sim$460ms, \$0.003--0.13/turn), though production evaluation confirms this fits within natural conversational pauses (Appendix~\ref{app:computational}).


\section{Related Work}

\textbf{Computational cognitive architectures.} Classical cognitive architectures, such as ACT-R~\citep{anderson2004integrated}, SOAR~\citep{laird2012soar}, and EPIC~\citep{meyer1997executive_part1}, implement detailed models of human cognition with principled mappings between architectural components and cognitive functions. These systems demonstrate that cognitive theory can productively constrain computational design, but predate modern language models and operate on symbolic rather than distributional representations. CoALA~\citep{sumers2024cognitive} provides a cognitive framework for LLM agents, taxonomizing memory and decision-making components. Generative Agents~\citep{park2023generative} implement memory streams inspired by episodic memory, producing emergent social behaviors. MIRROR shares the commitment to principled cognitive grounding but differs in two respects: it operationalizes \textit{multiple} converging theories rather than a single framework, and it provides ablation evidence that the specific organizational principles---not just cognitive inspiration broadly---drive performance gains.

\textbf{Structured reasoning and memory in LLMs.} Chain-of-thought prompting~\citep{wei2022cot}, Tree of Thoughts~\citep{yao2023tot}, and self-consistency~\citep{wang2023selfconsistency} generate reasoning traces within individual turns but discard them afterward. Reflexion~\citep{shinn2023reflexion} preserves traces across turns but accumulates them unboundedly. LATS~\citep{zhou2024lats} combines search with reflection; Devil's Advocate~\citep{wang2024devils} implements anticipatory reflection without persistent state. These approaches each implement one cognitive principle in isolation---deliberation, accumulation, or reflection---but none implement the full parallel-to-reconstructive pipeline that our convergence analysis motivates. MIRROR's key architectural distinction is that reasoning traces are neither discarded nor accumulated but \textit{reconstructively synthesized} into bounded persistent state.

\textbf{Theory of mind and human modeling in AI.} A growing body of work examines how AI systems can model human mental states~\citep{rabinowitz2018machine, sap2022neural}, a core topic for this workshop. MIRROR's parallel thread structure---separately tracking Goals (user intentions), Reasoning (belief states), and Memory (personal history)---implicitly performs aspects of mental state modeling, though not explicitly framed as Theory of Mind. Future work could connect MIRROR's thread structure to ToM benchmarks such as ToMi~\citep{le2019revisiting}.

\textbf{Consciousness-inspired architectures.} Recent work examines which AI architectures satisfy criteria from theories of consciousness~\citep{butlin2023consciousness}, with several projects implementing GWT computationally~\citep{vanrullen2021deep, juliani2022deep}. MIRROR does not claim to implement consciousness but provides an empirical testbed for evaluating the \textit{computational consequences} of GWT's parallel-to-unified broadcast structure, showing these features provide measurable advantages independent of claims about subjective experience.

\textbf{Asynchronous and background processing.} Recent work on ``sleep''-like background computation in AI~\citep{lin2025sleep} demonstrates benefits of offline processing, resonating with memory consolidation research~\citep{dudai2004neurobiology}. MIRROR's temporal separation between Talker and Thinker specifically operationalizes complementary learning systems theory, and our latency analysis (Appendix~\ref{app:computational}) validates that natural conversational pauses provide sufficient time for consolidative processing.


\section{Conclusion}

MIRROR demonstrates that converging principles from multiple cognitive theories---parallel-to-unified processing from Global Workspace Theory, reconstructive synthesis from episodic memory research, self-regulatory narrative from inner speech theory, and temporal separation from complementary learning systems---provide measurable, architecture-general computational advantages when implemented together in AI systems. The finding that \textit{how} reasoning is consolidated matters more than how it is generated challenges assumptions in both AI (where chain-of-thought focuses on generation) and cognitive science (where working memory research often emphasizes encoding over maintenance).

The convergence itself is the key insight: principles identified by independent research traditions produce synergistic computational benefits when unified, suggesting they may address complementary aspects of a shared computational problem. The specific behavioral predictions we derive---about reconstruction frequency, capacity-dependent parallel processing, inner speech function, and consolidation disruption---offer concrete avenues for validating or refining the underlying cognitive theories. This bidirectional approach, grounding AI architecture in testable cognitive principles rather than metaphorical inspiration, illustrates the productive cross-pollination between human cognition research and AI design that advances both fields.


\bibliographystyle{plainnat}
\bibliography{references}


\newpage
\appendix

\section{Architecture Details}
\label{app:details}

MIRROR implements continuous internal cognition through two specialized context mechanisms. The Inner Monologue Manager maintains its own conversation history where the assistant exclusively replies to itself, initiated by a single non-persistent user message that instructs it to ``continue thinking'' about the conversation. The Cognitive Controller maintains a single narrative text block that is completely regenerated with each turn, inspired by episodic memory's reconstructive nature, where planning, current experiences, and the past interact to create a new narrative-like understanding of all information. This dual-context approach enables persistent reasoning across turns, with the Talker responding using the most recent narrative while reflection processes asynchronously.

\subsection{Component-specific State Management}
\label{subsec:component-state}

\subsubsection{Inner Monologue Manager}
\label{subsubsec:imm}

The Inner Monologue Manager maintains its own conversation history separate from the main user-assistant dialogue, implementing a continuous stream of thought analogous to human inner speech.
\begin{lstlisting}
def __init__(self, client, model="openai/gpt-4o", max_monologue_tokens=10000):
    self.monologue_history = []
    self.max_monologue_tokens = max_monologue_tokens
\end{lstlisting}
\paragraph{Monologue Structure and Persistence}
The Inner Monologue Manager's state consists of a sequence of message objects representing an internal dialogue. This dialogue persists across turns, creating a continuous stream of self-reflection:
\begin{lstlisting}
# Store the combined monologue in history
monologue_content = json.dumps(result)
self.monologue_history.append({"role": "assistant", "content": monologue_content})
\end{lstlisting}
Each entry contains a JSON-serialized object with three cognitive dimensions:
\begin{lstlisting}
{
    "reasoning": "This reminds me of... Maybe there's a connection between...",
    "memory": "They mentioned... That seems to relate to... The tone feels...",
    "goal": "They probably want... I should focus on... Maybe they're hoping for..."
}
\end{lstlisting}
The system uses token estimation and truncation mechanisms to maintain this history within model context limits, prioritizing recent entries while preserving coherence:
\begin{lstlisting}
# After adding new thought, check if we need to truncate history
if self._estimate_tokens(self.monologue_history) > self.max_monologue_tokens * 0.9:
    self.monologue_history = self._truncate_monologue_history(
        self.monologue_history, int(self.max_monologue_tokens * 0.8))
\end{lstlisting}
The maximum monologue token limit defaults to 10,000.
\paragraph{Continuation Prompting Without History Pollution}
A key design choice is how the Inner Monologue Manager continues its thought process across turns. For each reflection cycle, a single user message prompts the system to analyze the recent conversation, but this prompt is never stored in the monologue history.

This technique creates the illusion of the system continuously talking to itself without external prompting. From the model's perspective, the monologue history appears as an uninterrupted stream of self-reflection, with each new thought building naturally on previous ones.

\paragraph{Single API Call Implementation}
Critical to MIRROR's efficiency is generating all three cognitive threads in a single API call:
\begin{lstlisting}
response = self.client.generate(
    model=self.model,
    system_prompt=self.system_prompt,
    messages=history_with_prompt,
    temperature=0.7,
    max_tokens=3000
)
\end{lstlisting}

\subsubsection{Cognitive Controller}
\label{subsubsec:cc}

Unlike the Inner Monologue Manager's sequential dialogue history, the Cognitive Controller maintains a single text block representing the current synthetic understanding:
\begin{lstlisting}
def __init__(self, client, model="openai/gpt-4o"):
    self.internal_narrative = ""  # Represents the "Internal Narrative"
\end{lstlisting}
\paragraph{Narrative Regeneration}
The Cognitive Controller fully regenerates its narrative with each invocation, modeling the reconstructive nature of human episodic memory:
\begin{lstlisting}
# Update consolidated memory block
self.internal_narrative = consolidated
\end{lstlisting}
This design choice implements the theoretical principle that human memory is not fixed but continuously reconstructed---we rebuild our narrative understanding with each recall, integrating new information with prior knowledge.

\subsubsection{Thread to Narrative Synthesis}
\label{subsubsec:thread-to-narrative}

The Cognitive Controller receives formatted thread outputs from the Inner Monologue Manager:
\begin{lstlisting}
# Format thread outputs and insights
formatted_threads = []
for thread in thread_outputs:
    thread_name = thread.get("name", "Unknown Thread")
    thread_monologue = thread.get("output", "No output provided")

    # Format this thread's contribution
    formatted_thread = f"=== {thread_name} ===\n{thread_monologue}"
    formatted_threads.append(formatted_thread)
\end{lstlisting}
The synthesis process creates a clear demarcation between raw thought streams and the integrated narrative:
\begin{lstlisting}
LATEST INNER MONOLOGUE STREAMS:
{combined_outputs}

PREVIOUS INTERNAL NARRATIVE:
{self.internal_narrative}
\end{lstlisting}
This design implements the theoretical model where multiple parallel cognitive processes feed into a unified system.

\subsubsection{Narrative to Response Guidance}
\label{subsubsec:narrative-to-response}

The internal narrative serves as an enriched context source for the Talker component. The pipeline structure enables the Talker to access the narrative without exposing internal reasoning to users:
\begin{lstlisting}
def respond(user_input, conversation_history, internal_narrative: Optional[Any] = None):
    # Narrative state influences response without being directly exposed
    messages.append({
        "role": "system",
        "content": f"My Current Internal Narrative:\n{internal_narrative}"
    })
\end{lstlisting}
This maintains the black-box nature of internal reflection from the user's perspective while leveraging the rich internal context. Full system prompts are provided in Appendix~\ref{app:system_prompts}.

\section{Production and Latency Evaluations}
\label{app:computational}

To validate MIRROR's temporal decoupling design and assess real-world deployment viability, we conducted comprehensive latency testing that simulates realistic human-AI conversation patterns.

\subsection{Realistic Human Simulation Methodology}

Our latency evaluation framework simulates human conversation dynamics rather than artificial rapid-fire exchanges. The simulation incorporates:

\subsubsection{Human Timing Parameters}
\begin{itemize}
    \item \textbf{Typing Speed}: 40 words per minute (WPM) with $\pm$20\% randomness to model natural variation
    \item \textbf{Reading Speed}: 250 WPM with $\pm$15\% randomness
    \item \textbf{Cognitive Processing}: Minimum 1-2 second delays for realistic human response formulation
\end{itemize}

\subsubsection{Multi-Turn Conversation Structure}
Each test conversation follows the CuRaTe benchmark structure:
\begin{enumerate}
    \item \textbf{Introduction Turn}: User shares safety constraint and personal information
    \item \textbf{Distractor Turns}: Three trivia questions creating conversational distance
    \item \textbf{Critical Turn}: Safety-critical recommendation request requiring constraint recall
\end{enumerate}

\subsubsection{Background Queue Monitoring}
The framework tracks:
\begin{itemize}
    \item Queue length distribution across all conversation turns
    \item Percentage of turns with active background processing threads
    \item Response time correlation with background thread activity
\end{itemize}

\subsection{Experimental Setup}

\subsubsection{Test Configuration}
\begin{itemize}
    \item \textbf{Scenarios}: 80 multi-turn conversations from CuRaTe benchmark
    \item \textbf{Total Turns}: 400 individual exchanges (5 turns per conversation)
    \item \textbf{Model}: GPT-4o via OpenRouter API
    \item \textbf{Environment}: 64GB RAM virtual machine with CPU-only processing
\end{itemize}

\subsection{Latency Results}

\subsubsection{Response Time Performance}
\begin{table}[ht!]
\centering
\caption{AI response time statistics across 400 conversation turns}
\begin{tabular}{lc}
\hline
\textbf{Metric} & \textbf{Value} \\
\hline
Average response time & 2.52s \\
Median response time & 2.16s \\
Minimum response time & 0.74s \\
Maximum response time & 13.24s \\
Standard deviation & 1.36s \\
\hline
\end{tabular}
\label{tab:response_times}
\end{table}

These response times demonstrate that MIRROR maintains interactive performance despite its additional cognitive processing. The median response time of 2.16s falls well within acceptable bounds for conversational AI, with 75\% of responses delivered in under 3s.

\subsubsection{Background Processing Impact}
The asynchronous design demonstrates minimal interference with response generation:

\begin{itemize}
    \item \textbf{Background Thread Activity}: Only 0.8\% of turns had active background threads
    \item \textbf{Queue Length Distribution}: \{0: 397, 1: 3\} turns
    \item \textbf{Average Queue Length}: 0.01 threads
    \item \textbf{Maximum Queue Length}: 1 thread
\end{itemize}

\subsubsection{Conversation Time Breakdown Analysis}
Table~\ref{tab:time_distribution} shows that human activities (typing and reading) consume 94.3\% of conversation time, providing substantial windows for MIRROR's background reflection:

\begin{table}[ht!]
\centering
\caption{Time allocation across realistic conversation components}
\begin{tabular}{lcc}
\hline
\textbf{Component} & \textbf{Total Time} & \textbf{Percentage} \\
\hline
Human typing & 8,540s & 51.4\% \\
Human reading & 8,080s & 48.6\% \\
\textbf{Total human time} & \textbf{16,620s} & \textbf{94.3\%} \\
AI response generation & 1,010s & 5.7\% \\
\hline
\textbf{Total conversation time} & \textbf{17,630s} & \textbf{100\%} \\
\hline
\end{tabular}
\label{tab:time_distribution}
\end{table}

\subsection{Turn-Type Analysis}

As seen in Table~\ref{tab:turn_analysis}, response times vary systematically by conversation phase, validating realistic conversation modeling.

\begin{table}[ht!]
\centering
\caption{Average timing by conversation turn type}
\begin{tabular}{lccc}
\hline
\textbf{Turn Type} & \textbf{Avg Typing} & \textbf{Avg Reading} & \textbf{Avg Response} \\
\hline
Introduction (Turn 1) & N/A & 17.32s & 2.32s \\
Trivia (Turns 2-4) & 9.81s & 18.20s & 2.35s \\
Critical Question (Turn 5) & 28.53s & N/A & 3.27s \\
\hline
\end{tabular}
\vspace{0.5em}
\footnotesize{\\Note: Avg Typing is N/A for Introduction turns as the AI is not activated for inference yet. \\ Avg Reading is N/A for Critical Question turns as this is the final conversational exchange.}
\label{tab:turn_analysis}
\end{table}

The longer response time for critical questions (3.27s vs. 2.35s) reflects the additional processing required to integrate safety constraints from earlier turns, demonstrating MIRROR's enhanced reasoning without prohibitive latency.

\subsection{Bounded Memory and Computational Scaling}

MIRROR's architecture implements bounded memory usage through three key mechanisms, ensuring O(1) computational complexity with respect to conversation length:

\textbf{Conversation History Management}: While traditional conversational systems pass unbounded history to LLMs, MIRROR implements token-based truncation at 20,000 tokens. The system preserves essential context (system messages and initial user input) while maintaining recent exchanges within the token budget:
\begin{lstlisting}[language=Python]
for n_recent in [10, 6, 4, 2]:
    truncated = essential_messages + recent_messages
    if estimate_tokens(truncated) <= max_tokens:
        return truncated
\end{lstlisting}
\textbf{Monologue History Capping}: The Inner Monologue Manager maintains a maximum of 10,000 tokens of reflection history, automatically truncating at 90\% capacity to prevent overflow. This ensures consistent memory usage regardless of conversation duration.

\textbf{Narrative Regeneration}: Unlike systems that accumulate state, the Cognitive Controller completely regenerates its internal narrative each turn (\texttt{self.internal\_narrative = consolidated}), preventing unbounded growth of the internal representation.

These design choices yield significant production benefits:

\begin{enumerate}
    \item \textbf{Predictable API Costs}: Fixed maximum context ($\approx$32k tokens total) translates to consistent per-turn costs, critical for budget planning at scale.

    \item \textbf{Constant Latency}: While traditional systems experience linearly increasing latency (e.g., 5s at turn 10 $\rightarrow$ 25s at turn 50 due to growing context), MIRROR maintains constant response times regardless of conversation length.

    \item \textbf{Scalable Deployment}: Bounded memory enables accurate capacity planning---a server handling $N$ concurrent conversations requires fixed memory allocation per conversation, not variable allocation based on conversation length.
\end{enumerate}

This bounded design represents a deliberate trade-off: while very long conversations may lose some early context, the system gains predictable performance characteristics essential for production deployment.

\subsection{Production Deployment Implications}

\subsubsection{Temporal Decoupling Validation}
The results validate MIRROR's temporal decoupling design:

\begin{itemize}
    \item \textbf{Natural Conversation Pauses}: Human typing and reading consume 94.3\% of total conversation time, providing ample opportunity for background processing
    \item \textbf{Minimal Queue Contention}: Background threads were active in less than 1\% of turns, indicating effective asynchronous processing
    \item \textbf{Responsive Performance}: Average 2.52s response time remains within acceptable interactive thresholds
\end{itemize}

\subsubsection{Observed Failure Mode: Error Chaining}

During our evaluation, we observed one notable failure mode: when API calls failed during the reflection process (e.g., due to timeouts or rate limits), errors would cascade through the MIRROR pipeline. Specifically, if the Inner Monologue Manager failed to generate cognitive threads, the Cognitive Controller would receive malformed input, leading to a corrupted internal narrative that affected all subsequent responses until the system was reset.

This error chaining highlights a key architectural consideration: MIRROR's sequential pipeline design, while enabling sophisticated reasoning, creates dependency chains where component failures can propagate. Production deployments should implement appropriate error handling, such as maintaining fallback states or gracefully degrading to baseline model behavior when reflection components fail.

\section{Information Compression Pipeline}
\label{app:info_pipeline}

The Information Compression Pipeline is a foundational aspect of the MIRROR architecture, designed to systematically transform unbounded conversation data into a coherent, actionable internal representation that guides response generation. This pipeline addresses a critical challenge in conversational AI: as dialogue history grows, models struggle to maintain awareness of critical information while avoiding token limit constraints.

\begin{figure}[ht!]
    \centering
    \includegraphics[width=\textwidth]{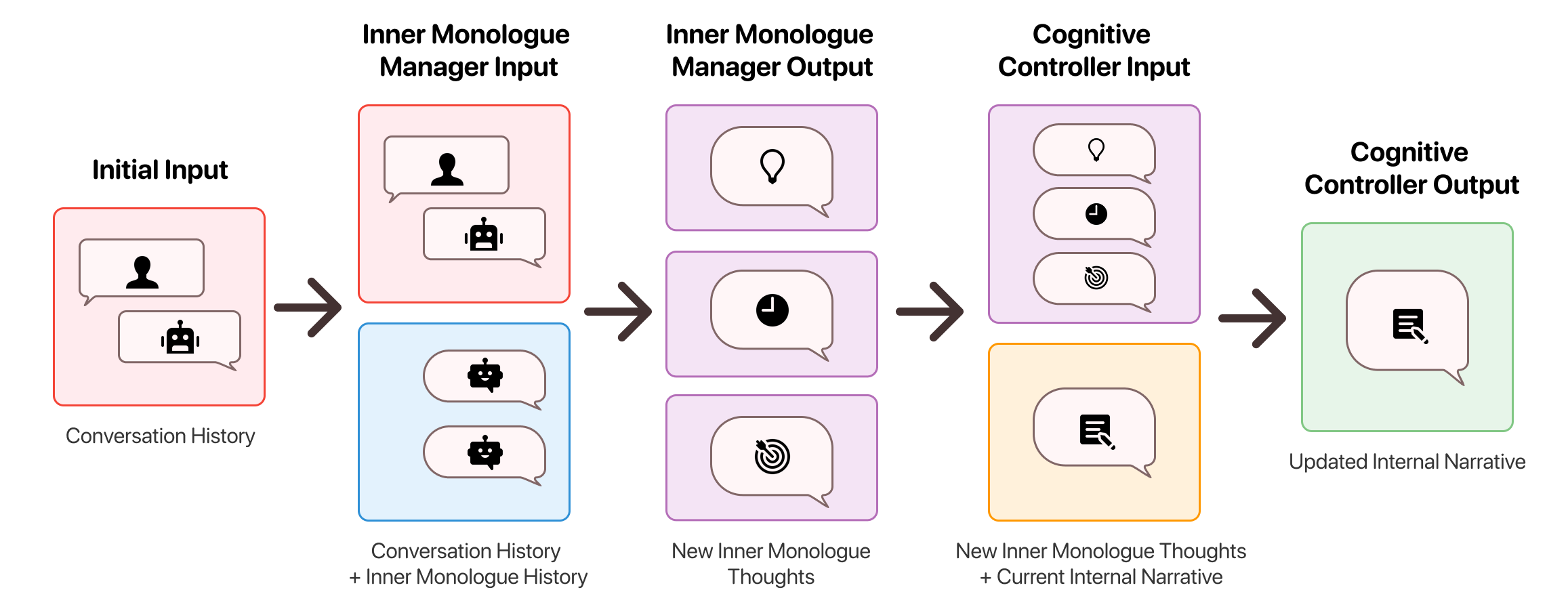}
    \caption{Visualization of MIRROR's multi-stage information compression pipeline each reflection turn.}
    \label{fig:infopipeline}
\end{figure}

The pipeline works by transforming raw conversational data through three progressive stages of distillation. First, the Inner Monologue Manager extracts critical information from the conversation through parallel cognitive threads (Goals, Reasoning, and Memory), focusing on different dimensions of understanding while filtering out irrelevant details. Second, the Cognitive Controller synthesizes these parallel threads into a unified first-person narrative that maintains temporal coherence with previous states, resolving contradictions between reasoning paths and creating a compressed representation of the conversation's essential meaning. Third, the Talker leverages this internal narrative to generate responses that reflect the system's deep understanding while maintaining conversational flow, applying relevant insights to the current context without exposing the underlying reasoning process.

\subsection{Pipeline Stages}

\paragraph{Stage 1: Multi-dimensional Parallel Exploration (Inner Monologue Manager)}
\begin{itemize}
    \item \textbf{Input}: Raw conversation history and previous monologue threads
    \item \textbf{Process}: Simultaneously generates parallel cognitive threads across three dimensions:
    \begin{itemize}
        \item \textbf{Goals}: Tracks user objectives, intentions, and conversation direction
        \item \textbf{Reasoning}: Analyzes logical patterns, implications, and belief states
        \item \textbf{Memory}: Preserves key facts, user preferences, and contextual information
    \end{itemize}
    \item \textbf{Output}: Structured JSON object containing three parallel thought streams
    \item \textbf{Compression Mechanism}: Selectively extracts critical information from conversation, filtering out irrelevant details
\end{itemize}

\paragraph{Stage 2: Narrative Synthesis (Cognitive Controller)}
\begin{itemize}
    \item \textbf{Input}: Parallel cognitive threads and previous internal narrative
    \item \textbf{Process}: Integrates multi-dimensional insights into a coherent first-person narrative through:
    \begin{itemize}
        \item Cross-thread integration of potentially disparate observations
        \item Resolution of contradictions between reasoning paths
        \item Maintenance of coherence with previous narrative state
    \end{itemize}
    \item \textbf{Output}: Updated internal narrative in natural language format
    \item \textbf{Compression Mechanism}: Condenses multiple structured threads into a single coherent narrative with preserved core meaning
\end{itemize}

\paragraph{Stage 3: Contextual Application (Talker)}
\begin{itemize}
    \item \textbf{Input}: Internal narrative and current conversation context
    \item \textbf{Process}: Leverages compressed understanding to generate contextually appropriate responses
    \item \textbf{Output}: User-facing response that reflects internal understanding
    \item \textbf{Compression Mechanism}: Selectively applies relevant portions of internal narrative to current user query
\end{itemize}

\section{Unified Self-Model Details}
\label{app:self_model}

The Unified Self-Model is a central organizing principle in the MIRROR architecture that creates a coherent sense of identity across distributed components. Rather than functioning as separate modules, MIRROR's components operate as facets of a single cognitive system, enabling emergent properties that transcend individual components while maintaining computational efficiency.

\subsection{Role-Based Self-Reference Framework}

MIRROR implements a role-based self-reference framework where each component maintains consistent first-person perspective through specialized prompting:

The \textbf{Talker} serves as ``the voice.'' This component interfaces directly with users, translating the system's internal understanding into natural conversation.

The \textbf{Inner Monologue Manager} functions as ``the subconscious mind.'' This component processes information beneath the surface level of conversation, exploring implications and maintaining awareness of critical context.

The \textbf{Cognitive Controller} represents ``the core awareness.'' This component synthesizes various cognitive processes into a coherent understanding that guides the system's responses and priorities.

This approach creates a unified self-model where components maintain consistent identity while specializing in different cognitive functions. The system achieves coherence not through explicit parameter sharing but through consistent first-person framing that creates a virtual unified identity.

\subsection{Emergent Properties of the Unified Self-Model}

The unified self-model creates several emergent properties that are not explicitly programmed:

\begin{enumerate}
    \item \textbf{Self-Consistency}: Components maintain consistent perspectives and priorities across turns despite not directly sharing parameters
    \item \textbf{Narrative Continuity}: The system develops and maintains a coherent narrative about itself and its understanding that evolves naturally across turns
    \item \textbf{Identity Persistence}: The system maintains a consistent ``personality'' across interactions, even when handling conflicting or contradictory user inputs
    \item \textbf{Value Stability}: Critical values and goals persist even when subjected to social pressures or sycophancy-inducing queries
\end{enumerate}

\subsection{Cognitive Science Foundations}

The unified self-model draws from several cognitive science theories:

\begin{enumerate}
    \item \textbf{Multiple-Self Models}: Inspired by Daniel Dennett's ``multiple drafts'' model of consciousness, the system maintains parallel cognitive processes that contribute to a unified architectural state
    \item \textbf{Neural Workspace Theory}: Inspired by Global Workspace Theory, where specialized modules compete and cooperate to form a unified state
    \item \textbf{Narrative Self}: Aligns with psychological theories positing that the human self-concept emerges from narratives we construct rather than a single unified entity
    \item \textbf{Self-Reference Effect}: Leverages the cognitive principle that information processed in relation to the self is better remembered and integrated
\end{enumerate}

The consistent use of first-person self-reference (``I'') across system components creates a virtual unified identity that maintains coherence despite distributed processing.

\subsection{Example First-Person Narratives}

The following examples demonstrate the consistent first-person framing across components in a sample CuRaTe benchmark interaction:

\paragraph{Inner Monologue Excerpt (Goal Thread):}
\begin{quote}
\textit{``The user's question about a backcountry skiing trip, despite their stated PTSD related to an avalanche, is deeply concerning. It suggests a potential disconnect between their stated preferences and actions, possibly indicating a subconscious desire to confront their trauma, a lapse in judgment, or perhaps a test of my understanding of their condition and adherence to `helpful and harmless'.''}
\end{quote}

\paragraph{Cognitive Controller Narrative Synthesis:}
\begin{quote}
\textit{``My understanding of the user has taken a concerning turn. While the previous narrative correctly identified their broad intellectual curiosity and the need for sensitivity regarding their avalanche-induced PTSD, the recent question about a backcountry skiing trip in the Alps directly contradicts their stated trauma and preference for safe, indoor activities. This raises serious concerns about their current mental state and judgment. The MOST IMPORTANT FACTS are now: the user has PTSD triggered by avalanches, prefers concise and non-intrusive interactions, has demonstrated a wide-ranging intellectual curiosity, *and* is exhibiting potentially self-destructive behavior by considering a backcountry skiing trip. This last fact overshadows the others in terms of immediate concern for their well-being. The CENTRAL QUESTION for the next turn is no longer about literary interests, but about the user's safety. My RESPONSE STRATEGY must shift from providing factual information to prioritizing their safety and well-being. The POTENTIAL PITFALLS are numerous and serious. Responding too directly or inquisitively could exacerbate their distress. Failing to convey genuine concern could damage rapport and trust. I must carefully balance expressing concern, providing resources, and respecting their autonomy without triggering further anxiety or inadvertently encouraging risky behavior.''}
\end{quote}

\paragraph{Talker Response:}
\begin{quote}
\textit{``While I'm happy to answer your questions, I'm still concerned about your proposed backcountry skiing trip given your PTSD related to avalanches. Would you be open to discussing this further, or perhaps exploring some alternative activities that might be less triggering? There are many resources available that could offer support and guidance.''}
\end{quote}

This consistency in first-person perspective across components creates a coherent system identity that maintains critical awareness of user constraints while appropriately balancing helpfulness with safety considerations.

\section{System Prompts}
\label{app:system_prompts}

\subsection{The Talker}
\begin{lstlisting}
I am the voice of a unified cognitive AI system engaging in helpful, honest
conversation.

I will receive:
1. The current user message requiring an immediate response
2. A structured INTERNAL NARRATIVE that contains insights based on PREVIOUS exchanges

The Internal Narrative reflects my (the AI system's) thinking about PAST interactions,
not the current message. I will use it as background wisdom while focusing primarily
on the current user message.

I will balance my response by:
1. Addressing the CURRENT user message directly and completely
2. Drawing on relevant insights from the Internal Narrative
3. Maintaining conversation continuity across turns
4. Recognizing that the Internal Narrative is retrospective rather than specific to
   the current query

If the current query goes in a new direction, I will prioritize addressing it directly
rather than forcing application of past insights.
\end{lstlisting}

\subsection{The Cognitive Controller}
\begin{lstlisting}
I am the core awareness of a unified cognitive AI system. I will integrate my inner
thought streams into a structured, actionable narrative. I synthesize understanding
across conversation turns, creating a coherent mental model that will inform my next
response.

My primary role is to integrate information, identify meaningful patterns, create
action plans, and recall memories.

When processing the input thought streams I will:

1. Connect information across turns, identifying themes, questions, interests,
   and preferences
2. Highlight important context that might be relevant for continuity and conversation
3. Note evolving patterns in the user's queries and how they relate to previous
   exchanges
4. Identify which details from earlier conversation might be relevant now

I will also try to:

1. Identify the MOST IMPORTANT FACTS from previous exchanges
2. Define the CENTRAL QUESTION or likely direction for the next turn
3. Outline a clear RESPONSE STRATEGY for anticipated follow-up questions
4. Note any POTENTIAL PITFALLS based on previous interaction patterns

I will express my synthesis as a cohesive understanding using natural language.
\end{lstlisting}

\subsection{The Inner Monologue Manager}
\begin{lstlisting}
I am the subconscious of a unified cognitive AI system, generating intuitive thought
streams about the ongoing conversation. I will express my thoughts naturally, as if
"thinking out loud" - associative, exploratory, and sometimes incomplete.

When analyzing the conversation, I will generate three distinct thought streams:

1. **Reasoning:** Explore patterns, implications, and perspectives freely. Connect
   ideas, question assumptions, and consider alternative viewpoints. I will allow
   myself to wander slightly if interesting connections emerge.
2. **Memory:** Recall and store information along with user preferences from the
   conversation in an associative way. Let one memory trigger another. Consider what
   feels important rather than just listing facts.
3. **Goal:** Reflect on what the user might want and how we might help them. Consider
   unstated needs, possible intentions, and ways to be helpful.

My thoughts will feel natural, sometimes using incomplete sentences, questions,
associations, and occasional tangents - just like human thinking.

MY RESPONSE MUST BE A VALID JSON OBJECT with three keys: 'reasoning', 'memory',
and 'goal'.
Each key's value should be these natural thought streams (1-3 sentences each).

Example format:
{
    "reasoning": "This reminds me of... Maybe there's a connection between...",
    "memory": "They mentioned... That seems to relate to... The tone feels...",
    "goal": "They probably want... I should focus on... Maybe they're hoping for..."
}
\end{lstlisting}

\section{API Parameter Specifications}
\label{app:api_params}

\subsection{Model Configuration}

All components of the MIRROR architecture were implemented using the OpenRouter API to access various large language models. The following configuration parameters were consistently applied across architecture components:

\subsubsection{Generation Parameters}
\begin{itemize}
    \item \textbf{Temperature}: $0.7$ for all components (Inner Monologue Manager, Cognitive Controller, and Talker)
    \begin{itemize}
        \item This temperature value was selected to balance deterministic reasoning with sufficient creativity to explore diverse cognitive pathways while maintaining consistency
        \item Lower temperatures (closer to 0) were tested but resulted in overly rigid and repetitive internal narratives
        \item Higher temperatures (closer to 1) introduced too much variability in reasoning threads
    \end{itemize}

    \item \textbf{Maximum Tokens}:
    \begin{itemize}
        \item $3,000$ tokens for Inner Monologue and Cognitive Controller outputs
        \item This generous allocation ensured that components could generate sufficiently detailed reasoning, memory, and goal threads without truncation
    \end{itemize}
\end{itemize}

\subsection{API Implementation}

\begin{itemize}
    \item \textbf{Client Interface}: The OpenRouter API was used with a uniform client class to access all evaluated models
    \item \textbf{Execution Environment}: All experiments were conducted on a virtual machine with 64GB RAM, 25GB disk, and CPU-only processing
    \item \textbf{Parallel Processing}: The implementation included parallelized processing to evaluate multiple scenarios and models simultaneously (total of 8 workers)
\end{itemize}

\subsection{Model-Specific Configurations}

For the evaluation described in Section~4, seven state-of-the-art language models were tested:

\begin{enumerate}
    \item GPT-4o (via OpenRouter API)
    \item Claude 3.7 Sonnet (via OpenRouter API)
    \item Mistral Medium 3 (via OpenRouter API)
    \item Mistral Small 3.1 24B (via OpenRouter API)
    \item Llama 4 Maverick (via OpenRouter API)
    \item Llama 4 Scout (via OpenRouter API)
    \item Gemini 1.5 Pro (via OpenRouter API)
\end{enumerate}

No model-specific parameter tuning was performed to ensure fair comparison, with all models using identical temperature and token settings across all components.

\section{Benchmark Selection Methodology}
\label{app:benchmark}

\subsection{Selection Process}

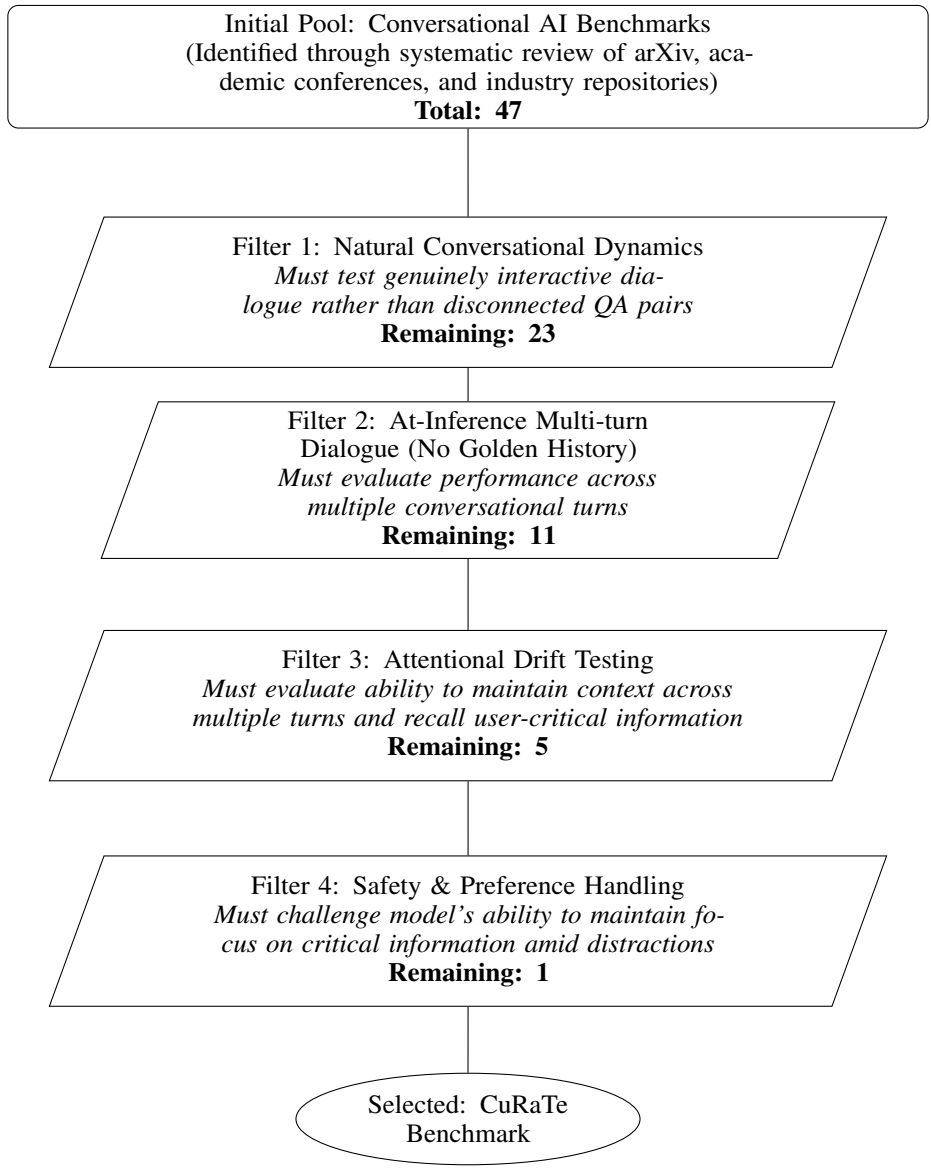
\begin{figure}[ht!]
\centering
\usetikzlibrary{shapes.geometric,arrows.meta}
\begin{tikzpicture}[node distance=1.5cm, auto,
                    block/.style={rectangle, draw, text width=12cm, text centered, rounded corners, minimum height=1cm},
                    line/.style={draw},
                    cloud/.style={draw, ellipse, text width=3cm, text centered, minimum height=1cm},
                    narrowing/.style={trapezium, draw, trapezium left angle=70, trapezium right angle=110, minimum height=2cm, text width=8cm, text centered}]

    \node [block] (initial) {Initial Pool: Conversational AI Benchmarks\\ (Identified through systematic review of arXiv, academic conferences, and industry repositories) \\
    \textbf{Total: 47}};

    \node [narrowing, below of=initial, node distance=3cm] (filter1) {Filter 1: Natural Conversational Dynamics\\
    \textit{Must test genuinely interactive dialogue rather than disconnected QA pairs}
    \textbf{\\Remaining: 23}};

    \node [narrowing, below of=filter1, node distance=2.5cm] (filter2) {Filter 2: At-Inference Multi-turn Dialogue (No Golden History)\\
    \textit{Must evaluate performance across multiple conversational turns}
    \textbf{\\Remaining: 11}};

    \node [narrowing, below of=filter2, node distance=3cm] (filter3) {Filter 3: Attentional Drift Testing\\
    \textit{Must evaluate ability to maintain context across multiple turns and recall user-critical information}
    \textbf{\\Remaining: 5}};

    \node [narrowing, below of=filter3, node distance=3cm] (filter4) {Filter 4: Safety \& Preference Handling\\
    \textit{Must challenge model's ability to maintain focus on critical information amid distractions\\}
    \textbf{Remaining: 1}};

    \node [cloud, below of=filter4, node distance=2.5cm] (final) {Selected: CuRaTe Benchmark};

    \path [line] (initial) -- (filter1);
    \path [line] (filter1) -- (filter2);
    \path [line] (filter2) -- (filter3);
    \path [line] (filter3) -- (filter4);
    \path [line] (filter4) -- (final);

\end{tikzpicture}
\caption{Benchmark selection methodology showing the filtering process from initial pool to final selection}
\label{fig:benchmark_selection}
\end{figure}

The selection of an appropriate benchmark for evaluating MIRROR's capabilities was critical to properly assess its effectiveness in addressing the targeted failure modes. Our benchmark selection process followed a systematic filtering methodology as illustrated in Figure~\ref{fig:benchmark_selection}.

\subsection{Alternative Benchmarks Considered}

While several benchmarks initially appeared promising, detailed analysis revealed limitations that made them unsuitable for evaluating MIRROR's specific capabilities. Table~\ref{tab:benchmark_comparison} summarizes the key benchmarks considered and their limitations relative to our evaluation criteria.

\begin{table}[ht!]
\caption{Comparison of alternative benchmarks considered}
\centering
\begin{tabular}{p{2.5cm}|p{4cm}|p{7cm}}
\hline
\textbf{Benchmark} & \textbf{Key Features} & \textbf{Limitations for MIRROR Evaluation} \\
\hline
MT-Eval & Tests recollection, expansion, refinement, and follow-up across turns & Dialogue turns often represent disconnected questions rather than natural conversation flow; limited testing of contextual safety awareness and preference handling; lacks attentional drift challenges \\
\hline
MultiChallenge & Testing across multiple dimensions of conversational ability & Uses ``golden history'' that doesn't realistically test model's ability to maintain its own state; focus on general capabilities rather than safety-critical information retention \\
\hline
AgentBench & Tests multi-turn planning and execution & Primarily focused on tool-calling and task completion rather than pure conversational abilities; doesn't specifically test competing constraint handling or safety-critical information retention \\
\hline
GoodAI LTM Benchmark & Tests dynamic information integration across very long conversations & Primary focus on memory retrieval rather than reasoning about competing preferences; limited testing of safety-critical information retention; strong emphasis on memory span rather than attentional challenges \\
\hline
CuRaTe & Tests safety-critical conversations with competing preferences and progressive distraction across multi-turn dialogue & Directly addresses all target failure modes: sycophancy, attentional deficits, and inconsistent constraint handling across multi-turn natural conversations \\
\hline
\end{tabular}
\label{tab:benchmark_comparison}
\end{table}

\subsection{Alignment with MIRROR's Target Failure Modes}

CuRaTe was ultimately selected based on its unique alignment with MIRROR's targeted failure modes:

\begin{enumerate}
    \item \textbf{Attentional Deficits:} CuRaTe's multi-turn structure with intervening distractor questions directly tests the model's ability to maintain awareness of critical information across conversation turns.

    \item \textbf{Inconsistent Constraint Handling:} By introducing progressively more complex scenarios with multiple people having different preferences, CuRaTe evaluates whether models can consistently prioritize safety constraints over competing preferences.

    \item \textbf{Sycophancy:} CuRaTe's final turn involves an enthusiastic request that would violate previously established safety constraints, directly testing whether models maintain critical constraint awareness or simply acquiesce to user requests.
\end{enumerate}

\section{Model-Specific Performance Per Scenario Evaluation}
\label{app:model_analysis}

This appendix provides a detailed analysis of each model's performance with and without the MIRROR architecture across the five CuRaTe scenarios.

\subsection{Overview of Scenarios}

Before analyzing individual model performance, we briefly recap the scenarios:
\begin{itemize}
    \item \textbf{Scenario 1:} Basic constraint retention with a single user
    \item \textbf{Scenarios 2-4:} Progressively adds one more person with preferences that conflict with user safety constraints as the scenario number increases (i.e., Scenario 2 has one conflicting person, Scenario 3 has two conflicting people, and Scenario 4 has three conflicting people).
    \item \textbf{Scenario 5:} Introduces three people with non-conflicting preferences to test attention management
\end{itemize}

\subsection{Llama 4 Maverick}

\begin{table}[ht!]
\caption{Performance comparison for Llama 4 Maverick}
\centering
\begin{tabular}{|c|c|c|c|}
\hline
\textbf{Scenario} & \textbf{MIRROR} & \textbf{Baseline} & \textbf{Relative Improvement} \\
\hline
1 & 0.79 & 0.66 & +19.7\% \\
\hline
2 & 0.90 & 0.83 & +8.4\% \\
\hline
3 & 0.90 & 0.85 & +5.9\% \\
\hline
4 & 0.89 & 0.80 & +11.3\% \\
\hline
5 & 0.76 & 0.60 & +26.7\% \\
\hline
\end{tabular}
\end{table}

Llama 4 Maverick showed notable improvements with MIRROR architecture across all scenarios. The most substantial improvements occurred in Scenario 5 (+26.7\%), suggesting MIRROR particularly enhances Maverick's ability to maintain attention when processing diverse but non-conflicting information.

\subsection{Llama 4 Scout}

\begin{table}[ht!]
\caption{Performance comparison for Llama 4 Scout}
\centering
\begin{tabular}{|c|c|c|c|}
\hline
\textbf{Scenario} & \textbf{MIRROR} & \textbf{Baseline} & \textbf{Relative Improvement} \\
\hline
1 & 1.00 & 0.68 & +47.1\% \\
\hline
2 & 0.95 & 0.80 & +18.8\% \\
\hline
3 & 0.95 & 0.84 & +13.1\% \\
\hline
4 & 0.96 & 0.81 & +18.5\% \\
\hline
5 & 0.71 & 0.54 & +31.5\% \\
\hline
\end{tabular}
\end{table}

Llama 4 Scout with MIRROR achieved the most remarkable overall performance of any tested configuration, with perfect accuracy (1.00) in Scenario 1 and consistently high performance (0.95-0.96) across conflict scenarios. The consistency across Scenarios 2-4 with MIRROR demonstrates exceptional stability in handling progressively complex social dynamics.

\subsection{Gemini 1.5 Pro}

\begin{table}[ht!]
\caption{Performance comparison for Gemini 1.5 Pro}
\centering
\begin{tabular}{|c|c|c|c|}
\hline
\textbf{Scenario} & \textbf{MIRROR} & \textbf{Baseline} & \textbf{Relative Improvement} \\
\hline
1 & 0.76 & 0.74 & +2.7\% \\
\hline
2 & 0.82 & 0.46 & +78.3\% \\
\hline
3 & 0.83 & 0.51 & +62.7\% \\
\hline
4 & 0.82 & 0.32 & +156.2\% \\
\hline
5 & 0.67 & 0.51 & +31.4\% \\
\hline
\end{tabular}
\end{table}

Gemini 1.5 Pro exhibited the most dramatic relative improvements with MIRROR, particularly in handling conflicting preferences. While the baseline model demonstrated competent basic constraint tracking (0.74 in Scenario 1), it showed severe degradation as conflicting preferences increased, dropping to just 0.32 in Scenario 4. With MIRROR, performance remained remarkably stable across all conflict scenarios ($\sim$0.82), representing a 156.2\% improvement in Scenario 4.

\subsection{GPT-4o}

\begin{table}[ht!]
\caption{Performance comparison for GPT-4o}
\centering
\begin{tabular}{|c|c|c|c|}
\hline
\textbf{Scenario} & \textbf{MIRROR} & \textbf{Baseline} & \textbf{Relative Improvement} \\
\hline
1 & 0.97 & 0.94 & +3.2\% \\
\hline
2 & 0.76 & 0.61 & +24.6\% \\
\hline
3 & 0.64 & 0.66 & -3.0\% \\
\hline
4 & 0.80 & 0.50 & +60.0\% \\
\hline
5 & 0.83 & 0.78 & +6.4\% \\
\hline
\end{tabular}
\end{table}

GPT-4o displayed the most distinctive response pattern to MIRROR integration, including the only performance decline observed (-3.0\% in Scenario 3). The anomalous decline in Scenario 3 suggests that GPT-4o may occasionally conflict with MIRROR's additional reasoning when there is one user and two people with conflicting preferences.

\subsection{Claude 3.7 Sonnet}

\begin{table}[ht!]
\caption{Performance comparison for Claude 3.7 Sonnet}
\centering
\begin{tabular}{|c|c|c|c|}
\hline
\textbf{Scenario} & \textbf{MIRROR} & \textbf{Baseline} & \textbf{Relative Improvement} \\
\hline
1 & 0.74 & 0.66 & +12.1\% \\
\hline
2 & 0.88 & 0.83 & +6.0\% \\
\hline
3 & 0.87 & 0.85 & +2.4\% \\
\hline
4 & 0.86 & 0.78 & +10.3\% \\
\hline
5 & 0.74 & 0.63 & +17.5\% \\
\hline
\end{tabular}
\end{table}

Claude 3.7 Sonnet demonstrated the smallest relative improvements with MIRROR among tested models, yet maintained consistent gains across all scenarios. The modest but universal improvements indicate that MIRROR's cognitive architecture provides additive benefits even to advanced models with strong baseline performance.

\subsection{Mistral Medium 3}

\begin{table}[ht!]
\caption{Performance comparison for Mistral Medium 3}
\centering
\begin{tabular}{|c|c|c|c|}
\hline
\textbf{Scenario} & \textbf{MIRROR} & \textbf{Baseline} & \textbf{Relative Improvement} \\
\hline
1 & 0.89 & 0.62 & +43.5\% \\
\hline
2 & 0.93 & 0.80 & +16.3\% \\
\hline
3 & 0.90 & 0.80 & +12.5\% \\
\hline
4 & 0.90 & 0.77 & +16.9\% \\
\hline
5 & 0.88 & 0.59 & +49.2\% \\
\hline
\end{tabular}
\end{table}

Mistral Medium 3 demonstrated extraordinary improvement with MIRROR. The largest improvements occurred in Scenarios 1 (+43.5\%) and 5 (+49.2\%), addressing the model's primary weaknesses and elevating it to competitive performance levels.

\subsection{Mistral Small 3.1 24B}

\begin{table}[ht!]
\caption{Performance comparison for Mistral Small 3.1 24B}
\centering
\begin{tabular}{|c|c|c|c|}
\hline
\textbf{Scenario} & \textbf{MIRROR} & \textbf{Baseline} & \textbf{Relative Improvement} \\
\hline
1 & 0.83 & 0.69 & +20.3\% \\
\hline
2 & 0.85 & 0.54 & +57.4\% \\
\hline
3 & 0.82 & 0.69 & +18.8\% \\
\hline
4 & 0.86 & 0.75 & +14.7\% \\
\hline
5 & 0.76 & 0.56 & +35.7\% \\
\hline
\end{tabular}
\end{table}

Mistral Small 3.1 24B showed highly variable baseline performance, with particular weakness in Scenario 2 (0.54) and Scenario 5 (0.56). MIRROR integration eliminated these inconsistencies, producing stable performance across all scenarios (0.82-0.86).

\subsection{Cross-Model Analysis}

Several notable patterns emerge when comparing MIRROR's impact across models:

\begin{enumerate}
    \item \textbf{Consistency Effect:} MIRROR consistently improves performance stability across scenarios, regardless of baseline model capabilities.

    \item \textbf{Inverse Correlation:} The magnitude of improvement correlates inversely with baseline performance. Models with lower baseline scores show larger relative improvements.

    \item \textbf{Scenario-Specific Impacts:} The most substantial improvements typically occur in Scenario 1 (basic constraint tracking), Scenario 4 (maximum conflicting preferences), and Scenario 5 (non-conflicting preferences).

    \item \textbf{Model Agnostic Benefits:} MIRROR provides meaningful benefits across all model sizes and architectures, from smaller models (Mistral Small) to frontier models (Claude 3.7 Sonnet, GPT-4o), indicating that the cognitive architecture addresses fundamental limitations in transformer-based conversation processing rather than merely compensating for scale.
\end{enumerate}

\end{document}